\documentclass[runningheads]{llncs}
\usepackage{array}
\usepackage{multirow}
\usepackage[T1]{fontenc}
\def\doi#1{\href{https://doi.org/\detokenize{#1}}{\url{https://doi.org/\detokenize{#1}}}}
\usepackage{graphicx}
\usepackage{mathrsfs}
\usepackage{amsfonts}
\usepackage{listings}
\usepackage{algorithm}
\usepackage{algorithmic}
\usepackage{bm}
\begin{document}
\title{FedGraph: an Aggregation Method from Graph Perspective  \thanks{Supported by organization x.}}
%
%
\author{Zhifang Deng\inst{1} \and
Xiaohong Huang\inst{1}\and
Dandan Li\inst{1}\and
Xueguang Yuan\inst{2}}
\authorrunning{Z. Deng et al.}
%
\institute{School of Computer Science (National Pilot Software Engineering School), Beijing University of Posts and Telecommunications, Beijing, China\and
School of Electronic Engineering, Beijing University of Posts and Telecommunications, Beijing, China\\}
\maketitle              
\begin{abstract}
	With the increasingly strengthened data privacy act and the difficult data centralization, Federated Learning (FL) has become an effective solution to collaboratively train the model while preserving each client's privacy. FedAvg is a standard aggregation algorithm that makes the proportion of dataset size of each client as aggregation weight. However, it can't deal with non-independent and identically distributed (non-i.i.d) data well because of its fixed aggregation weights and the neglect of data distribution. In this paper, we propose an aggregation strategy that can effectively deal with non-i.i.d dataset, namely FedGraph, which can adjust the aggregation weights adaptively according to the training condition of local models in whole training process. The FedGraph takes three factors into account from coarse to fine: the proportion of each local dataset size, the topology factor of model graphs, and the model weights. We calculate the gravitational force between local models by transforming the local models into topology graphs. The FedGraph can explore the internal correlation between local models better through the weighted combination of the proportion each local dataset, topology structure, and model weights. The proposed FedGraph has been applied to the MICCAI Federated Tumor Segmentation Challenge 2021 (FeTS) datasets, and the validation results show that our method surpasses the previous state-of-the-art by 2.76 mean Dice Similarity Score. The source code will be available at Github.
	
	\keywords{Federated Learning \and Brain Tumor Segmentation \and FedGraph \and Coarse-to-fine \and Topology.}
\end{abstract}
\section{Introduction}

With the increasing demand for the precise medical data analysis, deep learning methods are widely used in the medical image field. However, with the promulgation of the data act and the strengthening of data privacy, especially in the medical field, it has become more difficult to train models in large-scale centralized medical datasets. As one of the solutions, federated learning has attracted a lot of attention from researchers.

Federated learning(FL)[1][2] is a distributed machine learning paradigm in which all clients train a global model collaboratively while preserving their data locally. The naive repeat steps of FL are: (i) each client trains its model with local data; (ii) the server collects and aggregates the models from clients to get a global model, then delivers the global model to clients. The data flow between clients and server is the trained models rather than the original data, which avoids the leak of data privacy. As a crucial core of them, aggregation algorithm plays an important role in releasing data potential and improving global model performance. FedAvg[1], as pioneering work, is a simple and effective aggregation algorithm, which makes the proportions of local datasets size as the aggregation weights of local models. [3] proposed FedProx to limit the updates between local and global models by modifying the training loss of local models. FedMA[4] matches and averages the hidden elements with similar feature extraction signatures to construct the shared global model in a layer-wise manner. Federated learning has attracted the attention of scholars in more research fields. 

In medical image segmentation, Since [5] and [6] explored the feasibility of FL in brain tumor segmentation(BraTS), FL on medical image segmentation is in full swing. Liu et al. [7] proposed FedDG to make the model generalize to unseen target domains via episodic learning in continuous frequency space in retinal fundus image segmentation. Xia et al. [8] proposed Auto-FedAvg, where the aggregation weights are dynamically adjusted according to the data distribution, to accelerate the training process and get better performance in COVID-19
lesion segmentation. Zhang et al. [9] proposed SplitAVG to overcome the performance drops from data heterogeneity in FL by network split and feature map concatenation strategies in the BraTS task. More than this, the first computational competition on federated learning, Federated Tumor Segmentation(FeTS) Challenge\footnote[1]{https://fets-ai.github.io/Challenge/} [10] is held to measure the performance of different aggregation algorithms on glioma segmentation[11,12,13,14]. Leon et al. [15] proposed FedCostWAvg to get a notable improvement compared to FedAvg by including the cost function decreased during the last round and won the challenge. However, most of these methods only study the single granularity or add other regular terms to the aggregation method, without considering the finer granularity factors, which limit the performance of global model.

Different from the above methods, in this paper, we propose a novel aggregation strategy, FedGraph, which attempts to explore the aggregation algorithm of FL from the topological perspective of neural networks. After the server collects the local models, the FedGraph explores the internal correlations between local models by three aspects from coarse to fine: the proportion of each local dataset size, the topology structure of model graphs, and the model weights. The proportion of local dataset size factor is similar to FedAvg. We compute the topological correlation by mapping the local models into topological graphs. Meanwhile, the finer grain model weights correlations are taken into account. Through the weighted combination of three different granularity factors from coarse to fine, the proposed method promotes the more effective aggregation of local models. 

The primary contributions of this paper can be summarized as: 
(1) We propose FedGraph, a novel aggregation strategy which takes coarse-to-fine three factors: the sample size, the topology of model graphs and the model weights, especially from the topological perspective of neural network; 
(2) We propose an aggregation method which introduce the concept of graph into federated learning, and the aggregation weights can be adjusted adaptively;
(3) The superior performance is achieved by the proposed method, on the public FeTS challenge datasets.

\begin{figure}[t]
	\includegraphics[width=\textwidth]{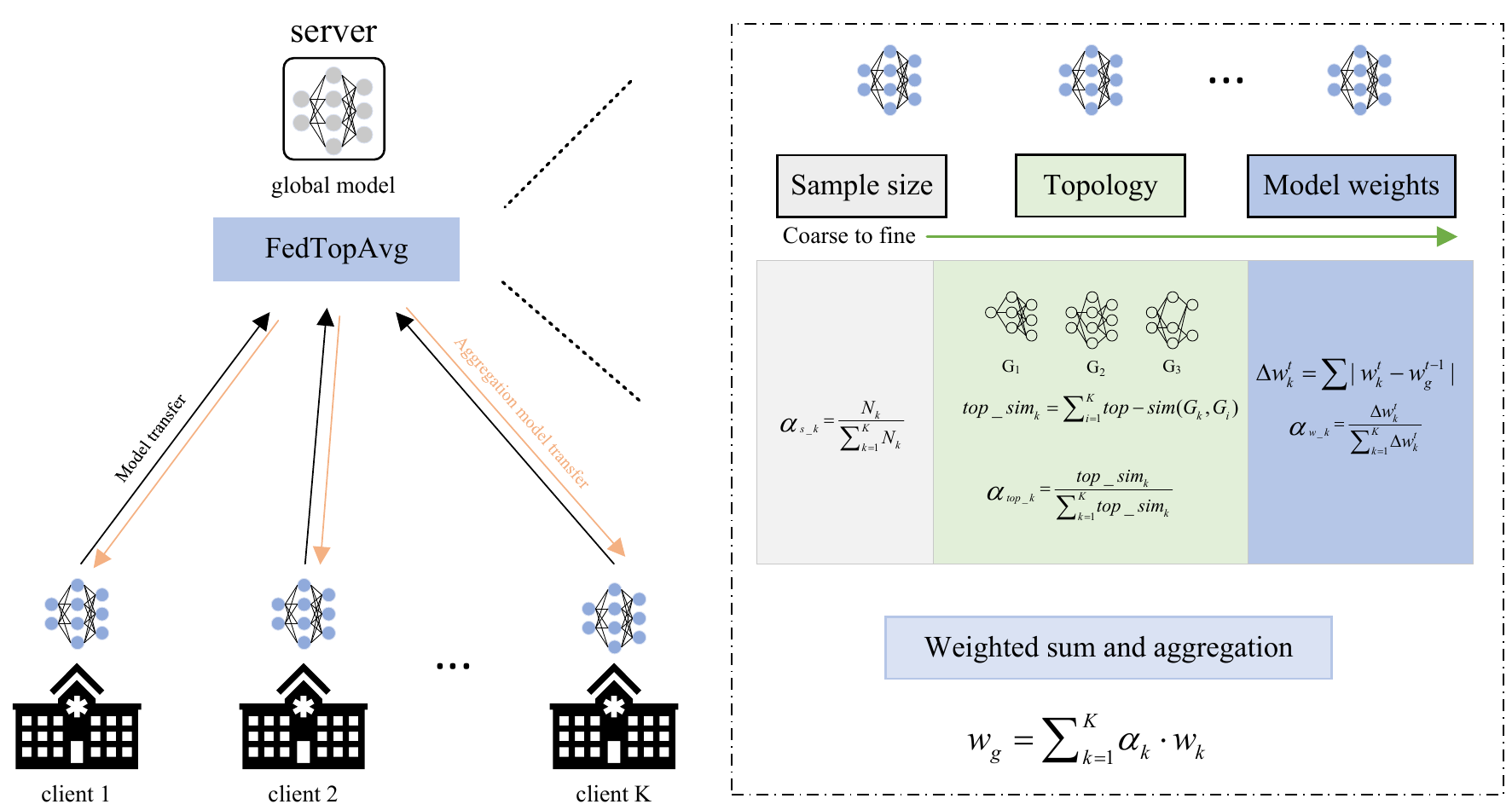}
	\caption{Overview of proposed FedGraph. The FedGraph aggregates local models and explore the correlations by considering coarse-to-fine three factors: the sample size of local datasets, network topology and model weights. } \label{fig1}
\end{figure}

\section{Related Work}
\textbf{Federated Learning} The FedAvg[1] plays a cornerstone role in federated learning tasks because of its efficient algorithm design, and the proportion of each client's dataset is taken as weight during the aggregation process. Recently, Some methods[19-24] based on FedAvg are proposed because of the poor performance caused by heterogeneous data distribution across clients in the real world. For example, FedProx[3] limits the bias between global model and local model by imposing additional regularization terms. FedDyn introduces a regularization term for local training based on the global model to address the problem. FedNova improves the aggregation stage by  normalizing and scaling the local updates of each client according to their local steps before updating the global model. SCAFFOLD addresses non-i.i.d problem by introducing variance among clients and employing the variance reduction technique. These methods have not been evaluated by real world datasets, although they have strict mathematical proof and good experimental results on manually created toy datasets. 

\textbf{Federated Learning in Medical Image Segmentation} Because the data act tightens data privacy and federal learning has the natural attribute of protecting data privacy, the researches of federated learning in the field of medical images have been impressive. [5] and [6] take the lead in discussing the application and safety of federated learning in brain tumor segmentation(BraTS). To solve the non-i.i.d challenges of FL in medical image field, FedDG[7] and FedMRCM[25] is proposed to address the domain shift issue between the source domain and the target domain, but the sharing of latent features may cause privacy concerns. Auto-FedRL[26] and Auto-FedAvg[8] is proposed to deal with the non-i.i.d problem by using an optimization algorithm to learn super parameters and aggregate weights. IDA[27] introduces the Inverse Distance of local models and the average model of all clients to handle non-i.i.d data. FedCostWAvg[15] overcome this issue to introducing the ratio of local loss of two adjacent rounds based on FedAvg as the aggregation weights of local models. While these methods study the single granularity or add other regular terms to the aggregation method alone, without considering the finer granularity factors.

\section{Method}

This section describes the overall pipline and the specific process of FedGraph first, and then we introduce the topology of neural networks. After that, we describe the details of FedGraph.

\subsection{Overview}
Suppose {\itshape K} clients with private data cooperate to train a global model and share the same neural network structure, 3D-Unet[16], which was provided by FeTS challenge and keep unchanged. For the clients, every client trains a local model {\itshape $w_i$} for local {\itshape E} epochs, and then delivers the local model to the server. The server aggregates local models to global model by computing the aggregation weights with the proposed FedGraph and assign to all clients. Repeat and until {\itshape T} rounds or other limits. An overview of the method is shown in Fig.1.

\subsection{FedGraph}
Clients send the updated local model back to the server each round. In round t, the global model {\itshape $w_g^t$} is aggregated by the server,

\begin{equation}
	\label{1}
	w_g^{t+1} = \sum_{k=1}^{K} \alpha_k \cdot w_k^{t}
\end{equation}

where $\alpha_k$ is the weight coefficient. In FedAvg, $\alpha_k= \frac{N_k}{N}, N=\sum_{k=1}^{K}N_k$, $N_k$ is the sample size of dataset in {\itshape $k$}-th local client. 

\begin{figure}[t]
	\includegraphics[width=\textwidth]{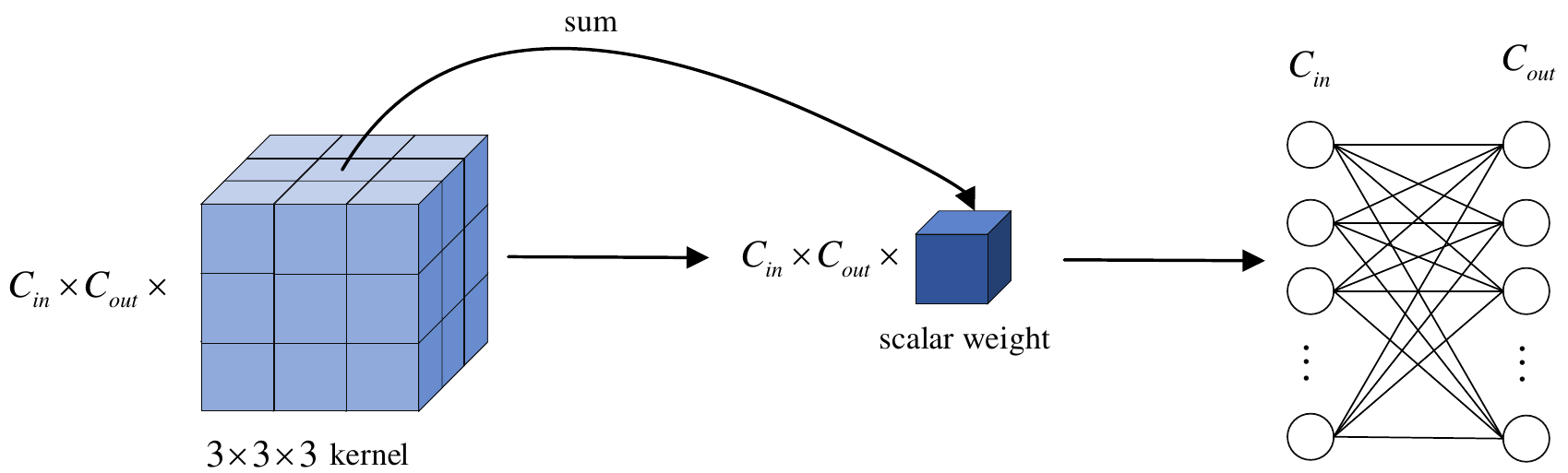}
	\caption{The process of graph mapping. } \label{fig2}
\end{figure}

In FedGraph, the computation of $\alpha_k$ goes through the following steps:

\textbf{Graph mapping}. Suppose the server has received local models trained by local data,  and we map them into the topological graph. Inspired by [17], take the {\itshape $j$}-th convolutional layer of {\itshape $k$}-th local model with 3D-Unet structure as an example, whose kernel dimension is {\itshape $3\times3\times3\times C_{in}\times C_{out}$}, it means this layer has $C_{in}\times C_{out}$ nodes with {\itshape $3\times3\times3$} filter, we can obtain $C_{in}\times C_{out}$ weight matrices of size {\itshape $3\times3\times3$}. Thus, we get  $C_{in}\times C_{out}$ nodes {\itshape $W$}$\in$$\mathbb{R}$$^{27}$. And then, we make every node {\itshape $W$} as scalar by averaging or summing, which can be formulated as:
\begin{equation}
	\label{2}
	w_{sum} = \sum_{d=0}^{2}\sum_{h=0}^{2}\sum_{w=0}^{2} W_{dhw}.
\end{equation}
It can be mapped into a graph whose structure is similar to the full connection layer after scalarization of convolutional layer. Given a {\itshape $3\times3\times3\times C_{in}\times C_{out}$} convolutional layer, the dimensions of its input and output are {\itshape $C_{in}$} and {\itshape $C_{out}$} respectively. So, we obtain a weight matrix  {\itshape $W_t$}$\in$$\mathbb{R}$$^{C_{in}\times C_{out}}$ after averaging or summing the weights of convolution kernel. We take the {\itshape $C_{in}$} and {\itshape $C_{out}$} as the number of node, and the weight summation $w_{sum}$ is the edge. The process of graphical model is shown in Fig. \ref{fig2}.

\textbf{Graph pruning}. The server collects local models from clients and makes the graph mapping on them to get {\itshape K} graphs which have same structure except the edge weights. To make the graphs more distinctive, the graph binarization is conducted. In detail, we differentiated these graphs by setting a threshold $\delta$, where the edge will be removed if the weight difference of each layer between the local models and global model in the last round is less than the threshold, otherwise, the edge will exist and the weight will be reset to 1. It can be simplified as:

\begin{equation}
	\label{3}
	edge= \left\{
	\begin{array}{l}
		1,  \quad \mid w_{kj}^t-w_{gj}^{t}\mid<\delta, \\  
		0,  \quad otherwise.
	\end{array}
	\right.
\end{equation}
\begin{equation}
	\label{4}
	\delta = Sort(\mid w_{kj}^t-w_{gj}^{t}\mid)[\lfloor \lambda \cdot l \rfloor ],  0 \leq  \lambda \leq 1.
\end{equation}
where in Eq. \ref{3}, 0 denotes the edge is removed and 1 denotes the edge exists and its weight is 1, $w_{kj}^t$ denotes edge weight of the $j$-th layer from the $k$-th graph in $t$-th round, also the weight summation of the $j$-th layer from the $k$-th local model in in $t$-th round, $w_{gj}^{t}$ is the the weight summation of the $j$-th layer from the global model in $t$-th round. The threshold $\delta$ varies adaptively with the weights of local models, and $\lambda$ is responsible for adjusting the degree of pruning. $Sort$ denotes that we sort weights in ascending order. After that we get $K$ discriminative graphs $G_i$, $i \in [1,K]$.

\textbf{Graph gravitational force}
In order to measure the degree of correlation between graphs and assign corresponding aggregation weights to local models, the further analysis is required on $K$ graphs obtained. In detail, we first measure the gravitational force between pairs of graphs to get the Graph Gravitational Force Matrix $\textbf{C} \in \mathbb{R}$$^{K\times K}$ by computing a matching between their sets of embeddings, where the Pyramid Match Graph Kernel[18] is employed. Then the Graph Gravitational Force Matrix $\textbf{C}$ is analyzed to compute the aggregation weights.  

For the completeness and readability of the article, we will briefly review the algorithm of Pyramid Match Graph Kernel[18]. Graph kernels is a powerful tool for graph comparison. The Pyramid Match Graph Kernel focuses on global properties of graph compared with other graph kernels based on local properties. The algorithm represent each graph as a bag-of-vectors, and then map this vector to multi-resolution histograms to campare the histograms with a weighted histogram intersection measure in order to find an approximate correspondence between the two sets of vectors. In detail, given a pair of graph  $G_1,G_2 \in \mathcal{G} $, $H_{G_1}^l$ and $H_{G_2}^l$ is the histograms of $G_1$ and $G_2$ at level $l$, and $H_{G_1}^l(i), H_{G_2}^l(i)$ is the number of vertices of $G_1, G_2$ that lie in the $i^{th}$ cell. The number of matched points in two sets is computed by the histogram intersection function:
\begin{equation}
	\label{5}
	I(H_{G_1}^l, H_{G_2}^l) = \sum_{i=1}^{D}min(H_{G_1}^l(i), H_{G_2}^l(i))
\end{equation}
The number of new matches at each level is $I(H_{G_1}^l, H_{G_2}^l)-I(H_{G_1}^{l+1}, H_{G_2}^{l+1})$ for $l=0,...L-1$, $L$ is the max level. The number of new matches found at each level in the pyramid is weighted according to the size of that level's cells, and the weight for level $l$ is $\frac{1}{2^{L-l}}$. The pyramid match kernel is defined as follows:
\begin{equation}
	\label{6}
	k_\triangle(G_1,G_2) = I(H_{G_1}^L,H_{G_2}^L)+\sum_{l=0}^{L-1}\frac{1}{2^{L-l}}(I(H_{G_1}^l,H_{G_2}^l))-I(H_{G_1}^{l+1},H_{G_2}^{l+1})
\end{equation}
more details can be found in [18].

\textbf{Graph weight}
According to the above method, the  Graph Gravitational Force Matrix $\bm{C}$ is obtained and $\bm{C}$ is a symmetric matrix, and the element $c_{kj}$ in matrix $\bm{C}$ denotes the gravitational force of $G_k$ and $G_j$, so the elements of the $k$-th row represents the gravitational force between $G_k$ and all graphs, so we can get the average gravitational force of $k$-th graph with all graphs:
\begin{equation}
	\label{7}
	c_k = \sum_{j=1}^{K}c_{kj}
\end{equation}
last, we normalize $c_k$ as the aggregation weight of the $k$-th local model, which we refer to as the topological weight:
\begin{equation}
	\label{8}
	\alpha_{top\_k} =  \frac{e^{c_k}}{\sum_{k=1}^{K} e^{c_k}} 
\end{equation}

\textbf{coarse-to-fine}
In our FedGraph, we take three factors from coarse to fine into consideration: the factor of sample size in local dataset $\alpha_s$, the factor of gravitational force between graphs $\alpha_{top}$ and the factor of model weights $\alpha_w$, where $\alpha_s$  is a rough adjustment of the aggregation weights of the overall model from the perspective of sample size, $\alpha_{top}$ is a moderate adjustment from the perspective of model topology, and $\alpha_w$ is a further detailed adjustment from the perspective of more refined model weights. Through the effective combination of them, more relevant information between local models is uncovered and more differentiated to promote the local models to be assigned more appropriate aggregation weights and aggregated a better global model. It can be denoted as:
\begin{equation}
	\label{9}
	\alpha_{w} = \frac{\frac{1}{\left| w_k-w_g \right|}}{\sum_{k=1}^{K}{\frac{1}{\left| w_k-w_g \right|}}}
\end{equation}
\begin{equation}
	\label{10}
	\alpha_k = \omega_s \cdot \alpha_s+\omega_{top} \cdot \alpha_{top}+\omega_w \cdot \alpha_w
\end{equation}
where $\omega_s, \omega_{top}, \omega_w$ are the weight coefficients of three factors: the factor of local dataset size proportion  $\alpha_s$, the factor of topology $\alpha_{top}$ and the factor of model weights $\alpha_w$, and $\omega_s+\omega_{top}+\omega_w=1$.

\begin{algorithm}[t]
	\caption{FedGraph}
	\label{alg1} 
	\begin{algorithmic}
		\REQUIRE
		{initial global model $w_g^0$, the global model weights in last round $w_g^{t-1}$, the local trained model weights $w_1^t,..., w_K^t$, and the local epochs $E$ in each client.}
		\ENSURE
		{The aggregation global model weights $w_g^T$.}  
		\FOR{$t=1 \to T$}
		\FOR{$k=1 \to K$ \textbf{in parallel} }
		\STATE $w_k^t \gets LocalTrain(k, w_g^{t-1})$
		\STATE $\triangleright$ upload $ w_k^t$ to the server
		\ENDFOR
		\STATE $\Delta w_k^t \gets \left| w_k^t-w_g^t \right| $
		\STATE $G_k^t \gets GraphMapping(\Delta w_k^t)$ 
		\STATE $p\_G_k^t \gets GraphPruning(G_k^t)$
		\STATE $c_{kj} \gets PyramidMatch(p\_G_k^t, p\_G_j^t)$
		\STATE $c_k \gets \sum_{j=1}^{K}c_{kj}$
		\STATE $\alpha_{top\_k}^t \gets  \frac{e^{c_k}}{\sum_{k=1}^{K} e^{c_k}} $,
		$\alpha_s^t \gets  \frac{N_k}{\sum_{k=1}^{K}N_k}$,
		$\alpha_{w}^t \gets \frac{\frac{1}{\Delta w_k^t}}{\sum_{k=1}^{K}{\frac{1}{\Delta w_k^t}}}$
		\STATE 	$\alpha_k^t = \omega_s \cdot \alpha_s^t+\omega_{top} \cdot \alpha_{top}^t+\omega_w \cdot \alpha_w^t$
		\STATE $w_g^{t+1} \gets \sum_{k=1}^{K} \alpha_k^t \cdot w_k^{t}$
		\ENDFOR 
		\RETURN{$w_g^T$}
		\STATE
		\STATE $LocalTrain(k, w_g^{t-1}):$
		\FOR{$t=1 \to E:$}
		\STATE Sample batch $x$ from client $k$'s training data
		\STATE Compute loss $loss(w;x)$
		\STATE Compute gradient of $w$ and update $w$
		\ENDFOR
		\RETURN $w$
		
	\end{algorithmic}
\end{algorithm}

\subsection{Algorithm}
We describe the algorithm of FedGraph in Algorithm \ref{alg1}. In each communicaition round $t$, the server send the global model $w_g^t$ to all clients, and then collects and aggragates the local models $w_1^t, ..., w_K^t$ with a set of weights $ \bm \alpha^t =[\alpha_1^t, ..., \alpha_K^t]$ to get an updated global model $w_g^{t+1}$ after the clients train and update their local models in parallel. The $\bm{\alpha^t}$ is determined by three factors $\alpha_s^t$, $\alpha_{top}^t$ in Eq. \ref{8} and $\alpha_w^t$ in Eq. \ref{9}, where the $\alpha_{top}^t$ is found by three steps: graph mapping in Eq. \ref{2}, graph pruning in Eq. \ref{3} and graph gravitational force in Eq. \ref{6}. After the weighted combination in Eq. \ref{10}, the server gets the new global model $w_g^t$ and deliver it to all clients to begin the next round.

\section{Experiments}
\subsection{Dateset and Settings}
The dataset used in experiments is provided by the FeTS Challenge organizer, which is the training set of the whole dataset about brain tumor segmentation. In order to evaluate the performance of FedGraph, we divide the dataset composed of 341 data samples into training set and validation set according to the ratio of 8:2, and the data is unevenly distributed between 17 data clients. 

The segmentation network, 3D-Unet, is provided by FeTS and kept unchanged, the learning rate is 1e-4 and the models train 10 epochs per round. Limited by the framework and official code mechanism, the total number of rounds of training is set to 70, although the performance of the algorithm does not converge to the best. The dice similarity coefficient and  hausdorff distance-95th percentile(HD95) are utilized to evaluate the performance of the global aggregated model. In the proposed FedGraph, the weight coefficients of three factors $\omega_s$, $\omega_{top}$, $\omega_w$ are  $0.4, 0.3, 0.3$ respectively.

\subsection{Results and Discussion} 
In this section, we compare our FedGraph with other state-of-the-art methods on the FeTS Challenge dataset.

\begin{table}[t]
	\begin{center}
	\caption{Comparison with other state-of-the-art methods on FeTS Chanllege dataset.}\label{tab1}
	\begin{tabular}{|m{2.5cm}|m{1.5cm}<{\centering} | m{1.5cm}<{\centering} | m{1.5cm}<{\centering}|m{1.5cm}<{\centering} | m{1.5cm}<{\centering} | m{1.5cm}<{\centering}|}
		\hline
		Method &DICE WT$\uparrow$ & DICE ET$\uparrow$& DICE TC$\uparrow$ &HD95 WT$\downarrow$ &HD95 ET$\downarrow$&HD95 TC$\downarrow$\\
		\hline
		FedAvg[1] & 90.91 &73.39 & 69.42 &3.96 & 40.99 & \textbf{15.37}\\
		FedCostWAvg[15] &90.98 &74.63  &69.46 &3.87 &33.76 &15.58\\
		FedGraph& \textbf{91.51} & \textbf{81.29} & \textbf{70.55} & \textbf{3.82} & \textbf{15.57} &16.10\\
		\hline
	\end{tabular}
\end{center}
\end{table}

\begin{figure}[t]
	\centering
	\begin{tabular}{m{4cm}<{\centering}m{4cm}<{\centering} m{4cm}<{\centering} }	
		\includegraphics[scale=0.3]{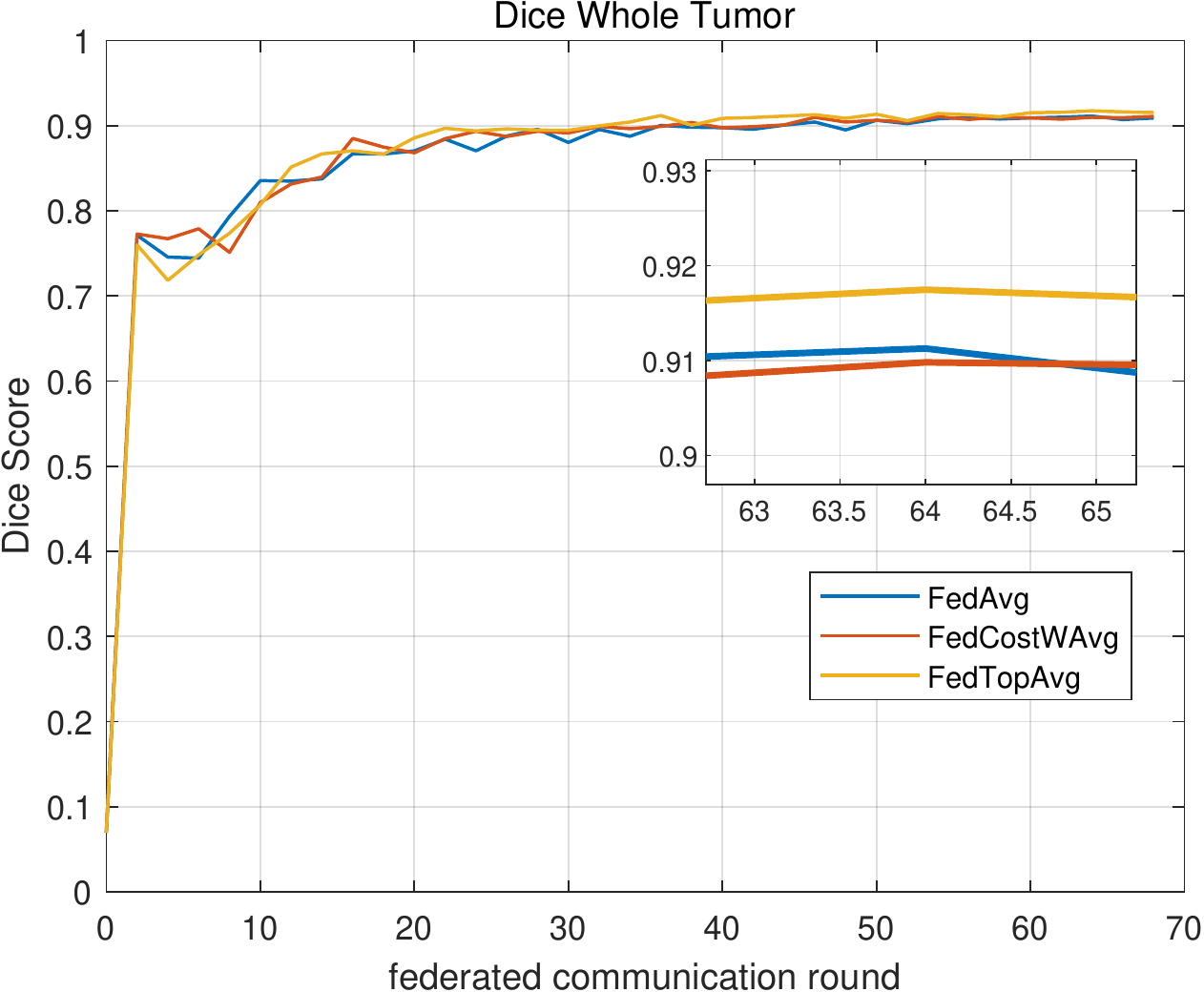}  & \includegraphics[scale=0.3]{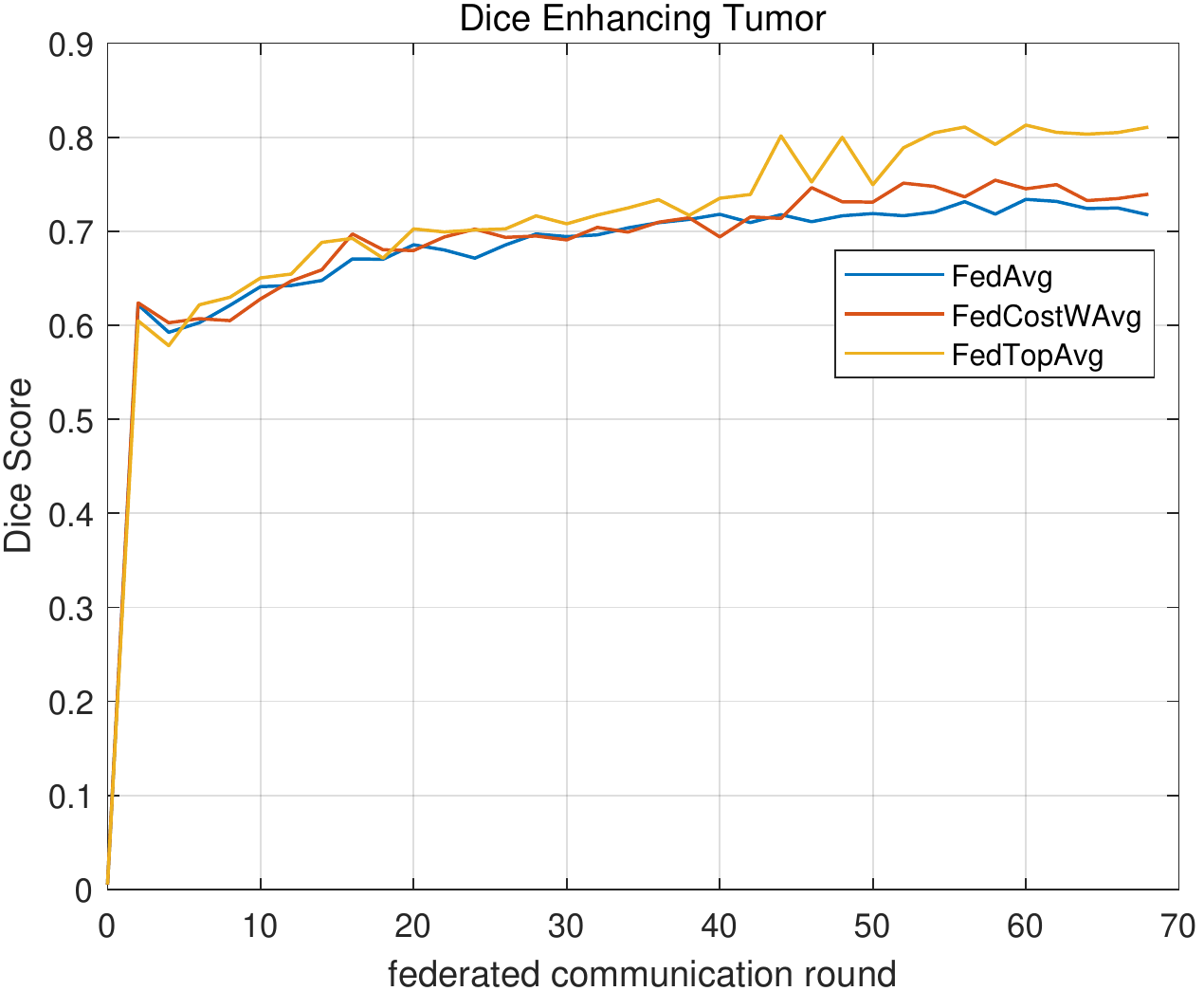} & \includegraphics[scale=0.3]{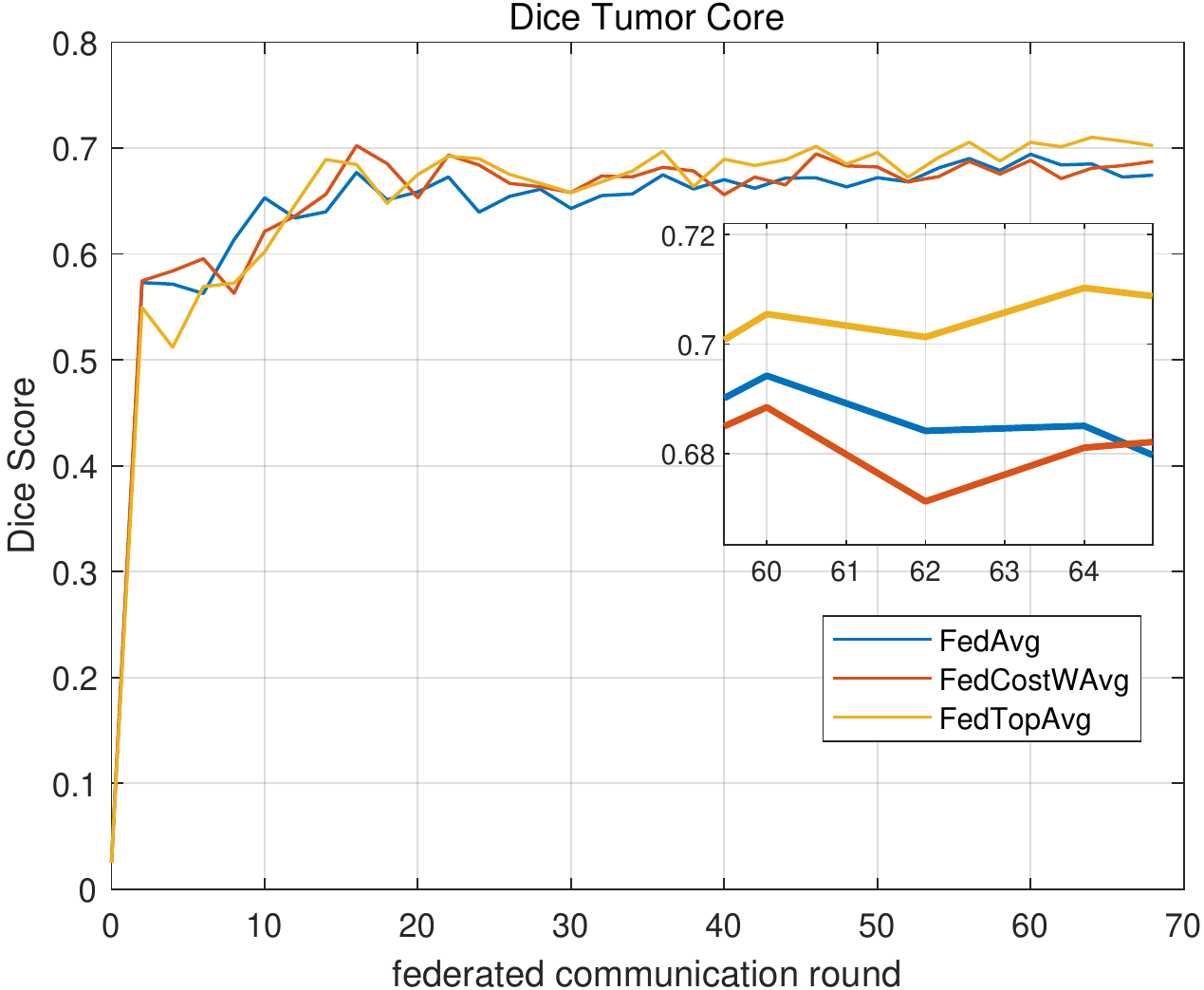} \\ 
		\includegraphics[scale=0.3]{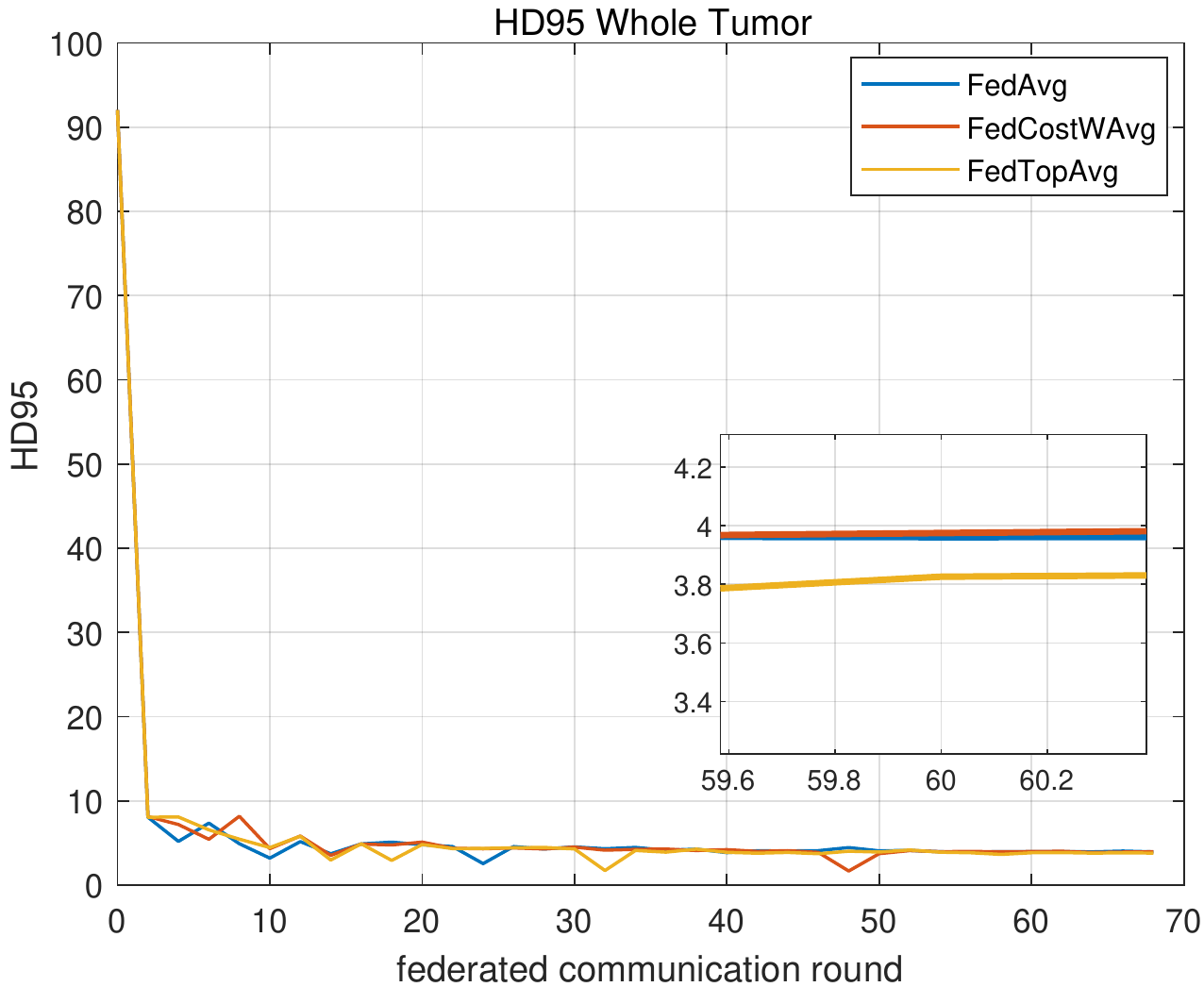}  & \includegraphics[scale=0.3]{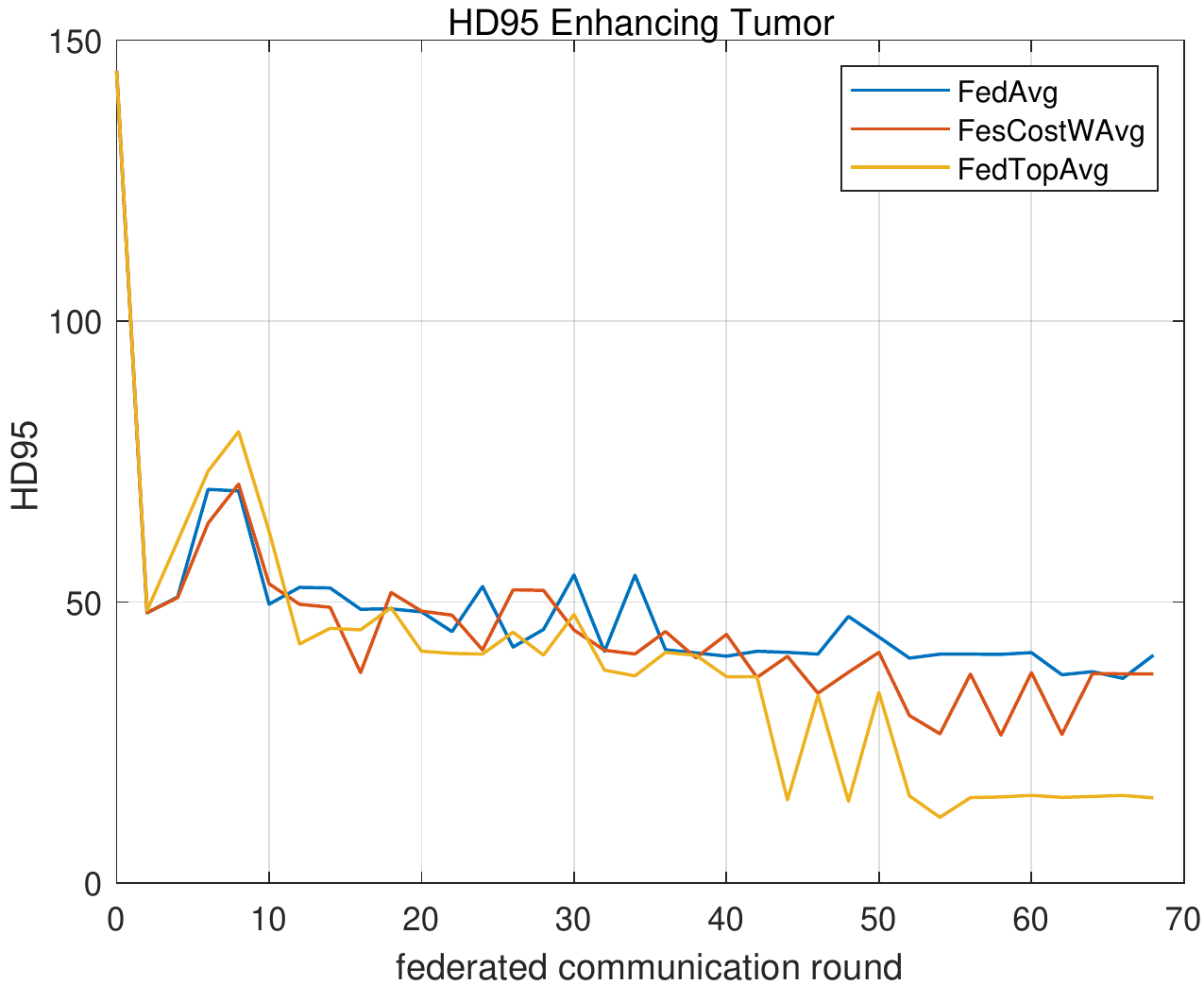} & \includegraphics[scale=0.3]{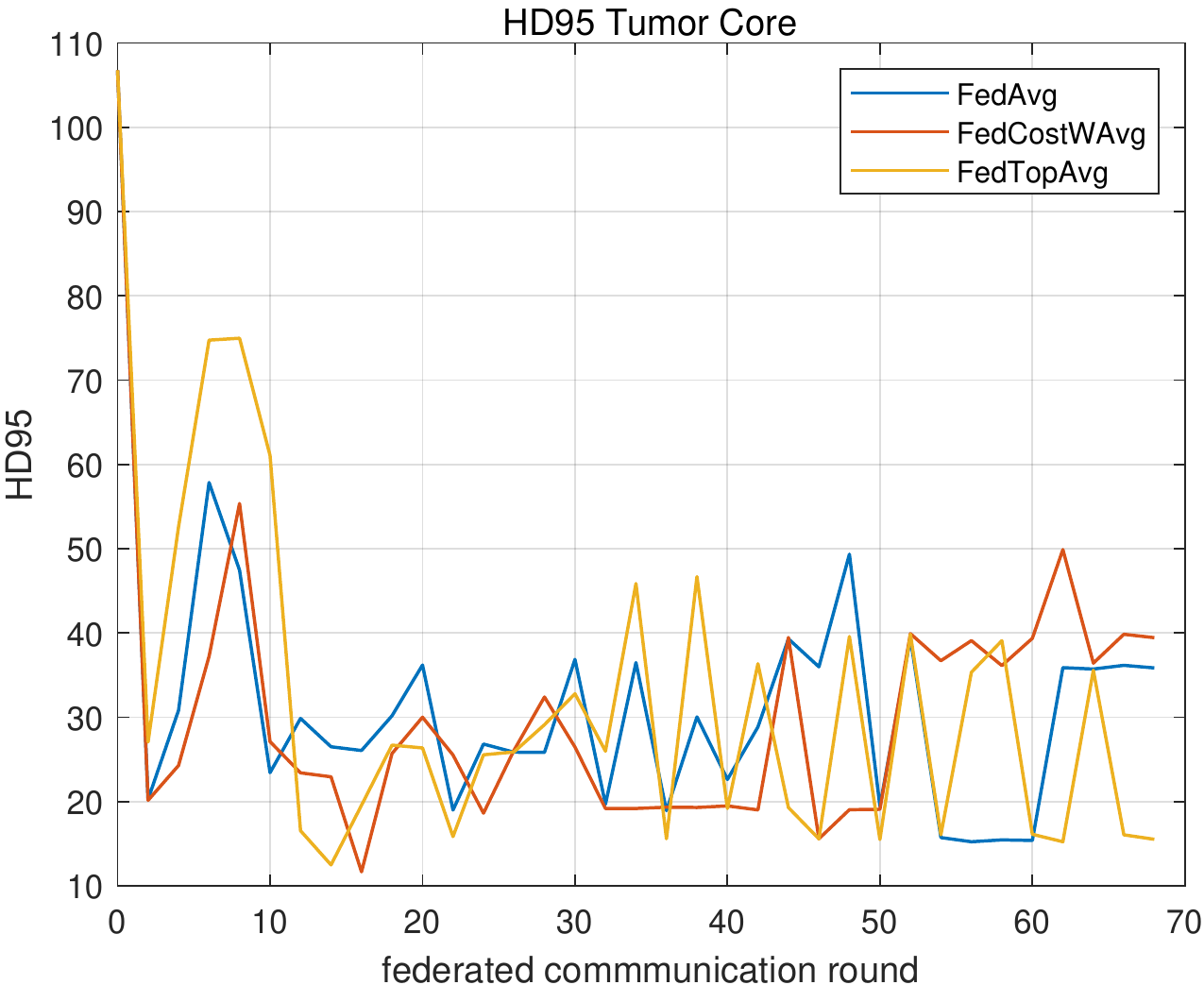} \\
	\end{tabular}
	\caption{Comparison of different factors and threshold in FedGraph. The sam\_top denotes the FedGraph with sample size factor and topology factor.} \label{fig:3}
\end{figure}

We evaluate the performance of our algorithm by comparing six indicators: the Dice Similarity Coefficient and Hausdorff Distance-95th percentile(HD95) of whole tumor(WT), enhancing tumor(ET), and tumor core(TC). As is shown in Table 1, we list the FedAvg, FedCostWAvg, the champion method of FeTS Challenge,  and the proposed FedGraph. Different from the original FedCostWAvg which changed the activation function of networks, our re-implemented version made the network unchanged to ensure a fair comparison. Through the quantitative comparison in Table 1, we can find that the proposed method FedGraph has achieved the best results in all indicators except the HD95 TC. Moreover, compared with FedCostWAvg, FedGraph has significantly improved the evaluation of segmentation performance, especially in enhancing tumor segmentation.

In addition, we counted the evaluation performance of these methods in the training process, as shown in Figure 3. We can observe that the FedGraph is in a leading position in almost all evaluation indicators after the training process tends to converge, which proved that the proposed method FedGraph can explore the correlations of local models better and achieved more excellent aggregation performance compared with other methods.

\begin{table}
	\begin{center}
		\caption{Comparison with other state-of-the-art methods on FeTS Chanllege dataset.}\label{tab1}
		\begin{tabular}{|m{2cm}|m{1cm}<{\centering}|m{1.2cm}<{\centering} |m{1.2cm}<{\centering} |m{1.2cm}<{\centering} |m{1.5cm}<{\centering} | m{1.5cm}<{\centering} | m{1.5cm}<{\centering}|}
			\hline
			Method &$\delta$ &sample & topology & weights & DICE WT & DICE ET & DICE TC \\
			\hline
			\multirow{5}{*}{FedGraph}
			&--&\checkmark &--&--& 90.91 &73.39 & 69.42 \\
			&0.001&\checkmark &\checkmark&--& 90.95 &77.47 & 69.23 \\
			&0.001&\checkmark &\checkmark&\checkmark& 91.42 &80.66 & 70.55 \\
			&0.01&\checkmark &\checkmark&\checkmark& \textbf{91.51} &\textbf{81.29} & 70.55 \\
			&0.1&\checkmark &\checkmark&\checkmark & 91.32 & 80.96 &\textbf{70.58}\\
			\hline
		\end{tabular}
	\end{center}
\end{table}

\subsection{Ablation Study}
To evaluate the effectiveness of each factor and the better configuration of FedGraph, we conduct the ablation study on the FeTS dataset and the results are shown in Table 2. First, we verified the impact of each factor. The FedGraph would degenerate into FedAvg when it contains only the proportion of local dataset size factor, and the topology factor can promote the performance of enhancing tumor segmentation effectively. With the gradual refinement of the granularity of the aggregation factor, the model weights factor cooperates the first two factors to make significant improvements of the test indicators, because the model weights factor is a finer granularity, and form effective complementarity with the first two factors to explore the more detailed relationship of the local models. Besides, we also test the different threshold $\delta$ to find the better configuration of FedGraph. The different values of $\delta$ denote the loose degree of topology, the smaller the $\delta$, the greater difference of topology. We can observe that it is a suitable degree when $\delta=0.01$ in our task, looser or stricter will lead to performance degradation. Fig.4 shows the comparison of dice in WT, ET, and TC for more intuitive representation.

\begin{figure}
	\centering
	\begin{tabular}{m{4cm}<{\centering}m{4cm}<{\centering} m{4cm}<{\centering} }	
		\includegraphics[scale=0.3]{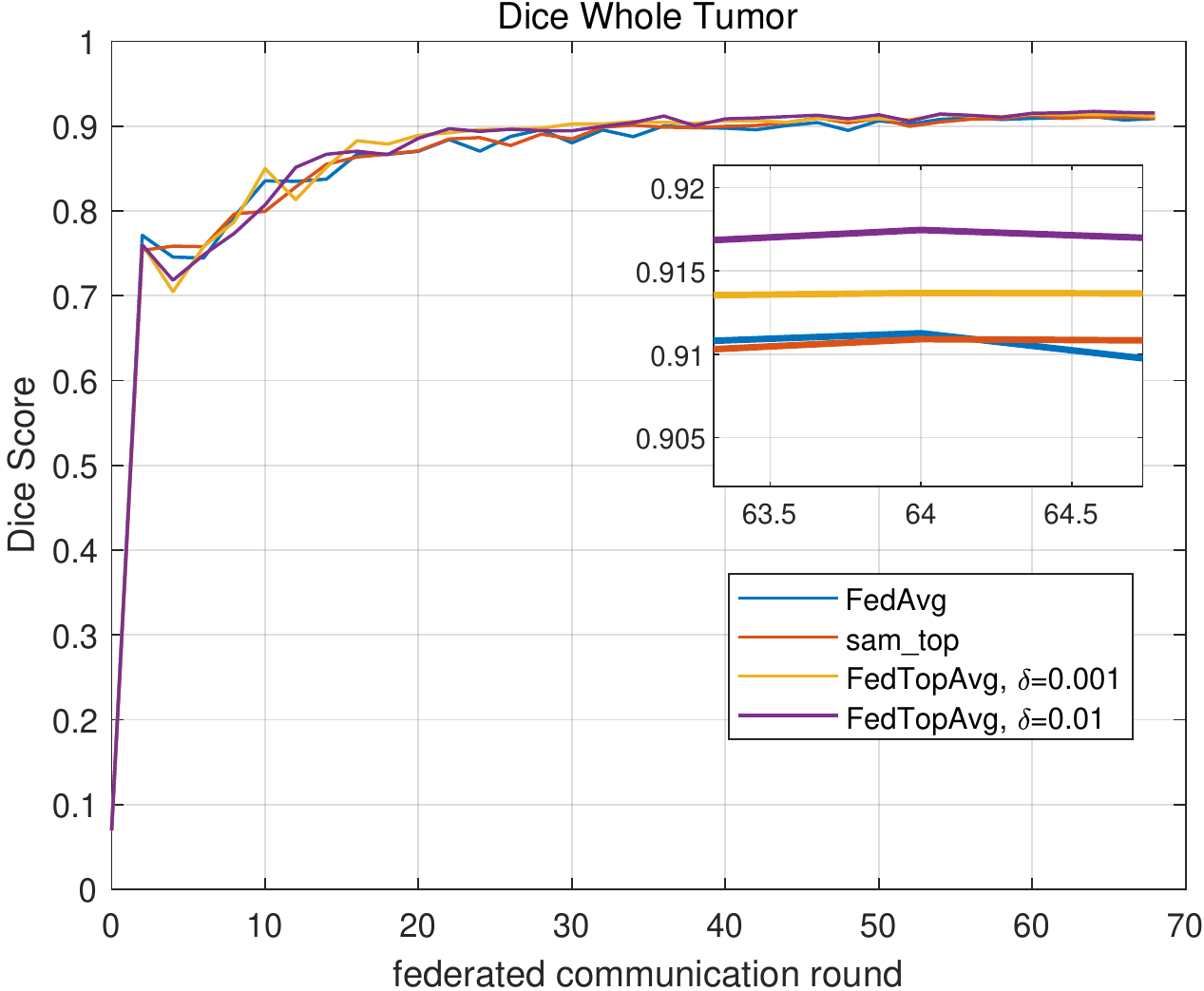}  & \includegraphics[scale=0.3]{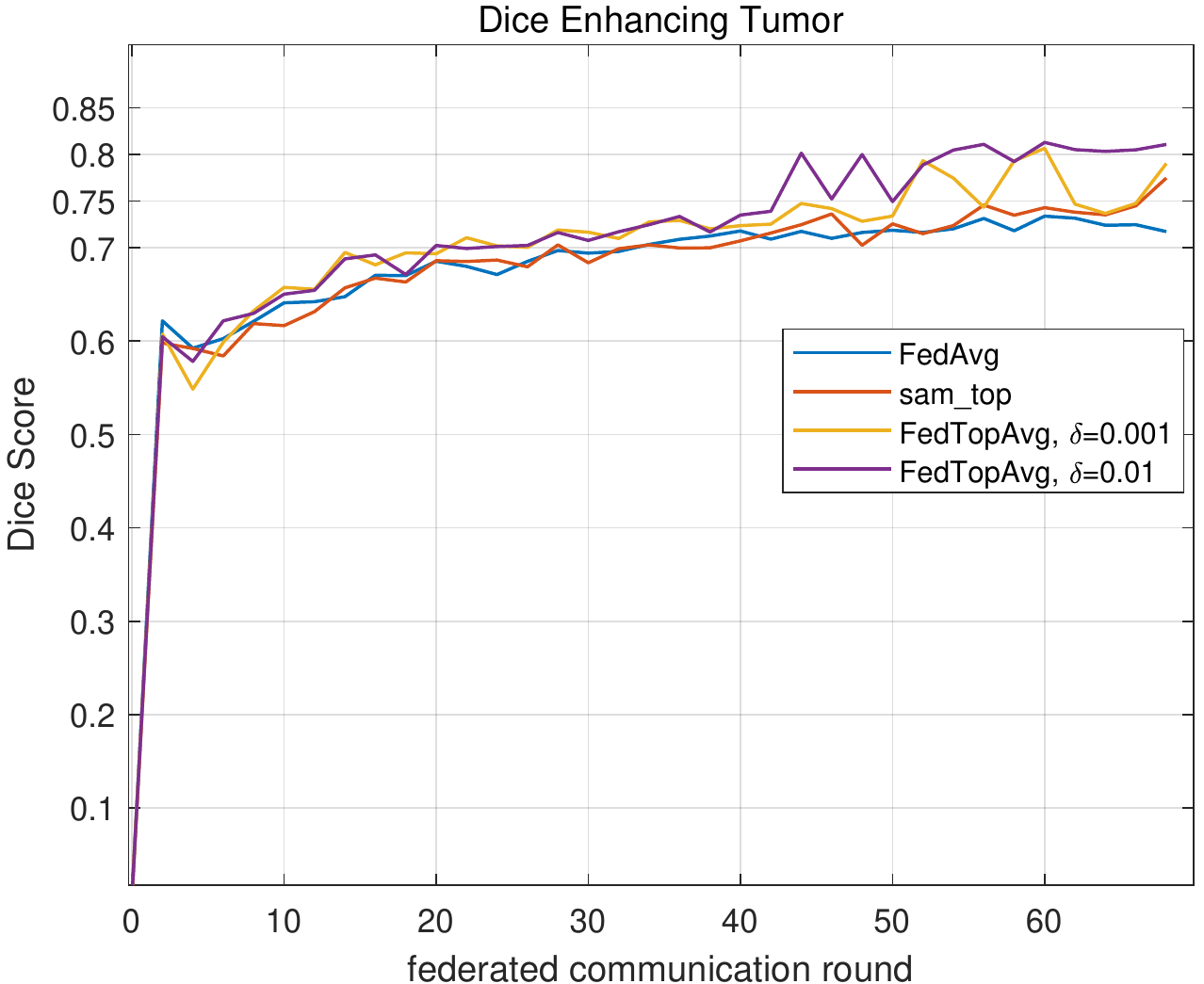} & \includegraphics[scale=0.3]{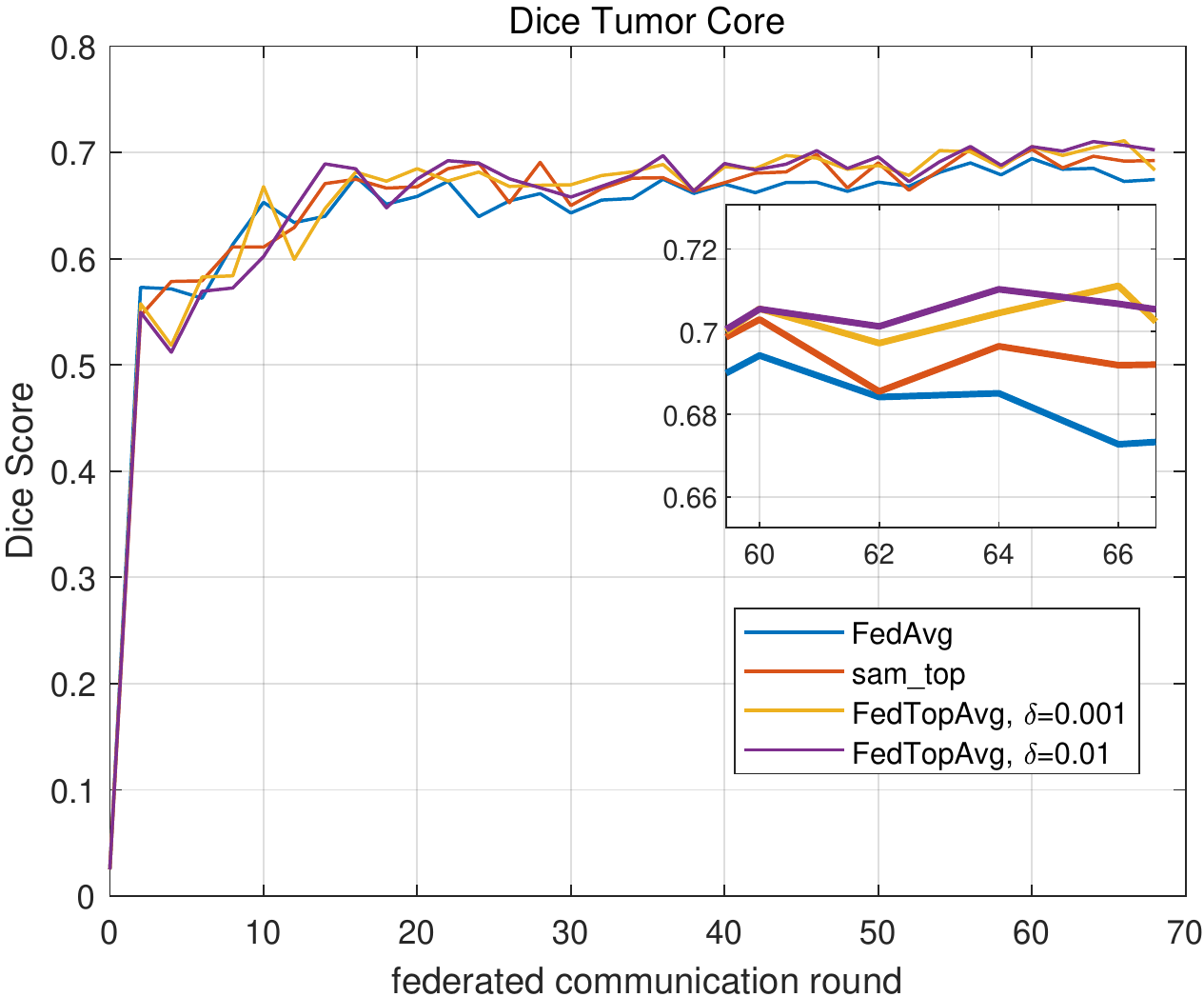} 
	\end{tabular}
	\caption{Comparison of different factors and threshold in FedGraph. The sam\_top denotes the FedGraph with sample size factor and topology factor.} \label{fig:3}
\end{figure}

\begin{table}
	\begin{center}
		\caption{Comparison with other state-of-the-art methods on FeTS Chanllege dataset.}\label{tab1}
		\begin{tabular}{|m{1.7cm}|m{1cm}<{\centering}|m{1.5cm}<{\centering} |m{1.7cm}<{\centering}|m{1.5cm}<{\centering} | m{1.5cm}<{\centering} | m{1.5cm}<{\centering}|}
			\hline
			Method &$\delta$ &similarity & dissimilarity  & DICE WT & DICE ET & DICE TC \\
			\hline
			
			FedAvg &--&-- & -- & 90.91 &73.39 & 69.42 \\
			\hline
			\multirow{3}{*}{FedGraph}
			&0.001&\checkmark &--& 91.20 &75.74 & \textbf{70.90} \\
			&0.001&-- &\checkmark& \textbf{91.42} &\textbf{80.66} & 70.55 \\
			
			\hline
		\end{tabular}
	\end{center}
\end{table}

Besides, in order to further explore the impact of local model gravitational force aggregation and repulsive force aggregation on the final global model in the aggregation process, we conduct ablation experiments under the condition of $\delta=0.001$, and the results are shown in table 3. Both  gravitational force aggregation and repulsive force aggregation methods show better results than FedAvg, and the repulsive force aggregation method is more suitable for the non-i.i.d datasets in our experiment. We will continue to study the aggregation method of the combination of the two in our future work.

\section{Conclusion}
In this paper, we introduced FedGraph, which improves the local model aggregation by taking coarse-to-fine three factors into account: the proportion of each local dataset size, the topology factor of graphic models, and the model weights. It can explore the correlations of local models better by an appropriately weighted combination of these factors. The further experiment results on MICCAI Federated Tumor Segmentation Challenge (FeTS) dataset demonstrate the superiority of the FedGraph.

\subsubsection{Acknowledgements} Please place your acknowledgments at
the end of the paper, preceded by an unnumbered run-in heading (i.e.
3rd-level heading).

%
%
%

\clearpage
\begin{thebibliography}{8}
	\bibitem{ref_article1}
	Brendan McMahan, Eider Moore, Daniel Ramage, Seth Hampson, and Blaise Aguera y Arcas. Communication-efficient learning of deep networks from decentralized data. In Artificial Intelligence and Statistics, pages 1273-1282. PMLR, 2017. 1, 2, 3
	
	\bibitem{ref_article1}
	Qiang Yang, Yang Liu, Tianjian Chen, and Yongxin Tong. Federated machine learning: Concept and applications. ACM Transactions on Intelligent Systems and Technology (TIST), 10(2):1-19, 2019. 1
	
	\bibitem{ref_article1}
	Sahu, A.K., Li, T., Sanjabi, M., Zaheer, M., Talwalkar, A., Smith, V.: On the
	convergence of federated optimization in heterogeneous networks. arXiv preprint arXiv:1812.06127 3 (2018) 3
	
	\bibitem{ref_article1}
	Hongyi Wang, Mikhail Yurochkin, Yuekai Sun, Dimitris Papailiopoulos, and Yasaman Khazaeni. Federated learning with matched averaging. ICLR, 2020. 2, 5
	
	\bibitem{ref_article1}
	Sheller, M.J., Reina, G.A., Edwards, B., Martin, J., Bakas, S.: Multi-institutional deep learning modeling without sharing patient data: A feasibility study on brain tumor segmentation. In: MICCAI Brainlesion Workshop. pp. 92{104 (2018)
		
		\bibitem{ref_article1}
		Li W, Milletarì F, Xu D, et al. Privacy-preserving federated brain tumour segmentation[C]//International workshop on machine learning in medical imaging. Springer, Cham, 2019: 133-141.
		
		\bibitem{ref_article1}
		Liu Q, Chen C, Qin J, et al. Feddg: Federated domain generalization on medical image segmentation via episodic learning in continuous frequency space[C]//Proceedings of the IEEE/CVF Conference on Computer Vision and Pattern Recognition. 2021: 1013-1023.
		
		\bibitem{ref_article1}
		Xia Y, Yang D, Li W, et al. Auto-FedAvg: Learnable Federated Averaging for Multi-Institutional Medical Image Segmentation[J]. arXiv preprint arXiv:2104.10195, 2021.
		
		\bibitem{ref_article1}
		Zhang M, Qu L, Singh P, et al. SplitAVG: A heterogeneity-aware federated deep learning method for medical imaging[J]. arXiv preprint arXiv:2107.02375, 2021.
		
		\bibitem{ref_article1}
		Pati, S., Baid, U., Zenk, M., Edwards, B., Sheller, M., Reina, G.A., Foley, P., Gruzdev, A., Martin, J., Albarqouni, S., et al.: The federated tumor segmentation (fets) challenge. arXiv preprint arXiv:2105.05874 (2021)
		
		\bibitem{ref_article1}
		Bakas, S., Akbari, H., Sotiras, A., Bilello, M., Rozycki, M., Kirby, J.S., Freymann, J.B., Farahani, K., Davatzikos, C.: Advancing the cancer genome atlas glioma mri collections with expert segmentation labels and radiomic features. Scientic data4(1) (2017) 1-13
		
		\bibitem{ref_article1}
		Reina, G.A., Gruzdev, A., Foley, P., Perepelkina, O., Sharma, M., Davidyuk, I., Trushkin, I., Radionov, M., Mokrov, A., Agapov, D., et al.: Open: An open-source framework for federated learning. arXiv preprint arXiv:2105.06413 (2021)
		
		\bibitem{ref_article1}
		Sheller, M.J., Edwards, B., Reina, G.A., Martin, J., Pati, S., Kotrotsou, A.,
		Milchenko, M., Xu, W., Marcus, D., Colen, R.R., et al.: Federated learning in
		medicine: facilitating multi-institutional collaborations without sharing patient data. Scientific reports 10(1) (2020) 1-12
		
		\bibitem{ref_article1}
		Koer, F., Berger, C., Waldmannstetter, D., Lipkova, J., Ezhov, I., Tetteh, G.,
		Kirschke, J., Zimmer, C., Wiestler, B., Menze, B.H.: Brats toolkit: translating brats brain tumor segmentation algorithms into clinical and scientific practice. Frontiers in neuroscience 14 (2020) 125
		
		\bibitem{ref_article1}
		Mächler L, Ezhov I, Kofler F, et al. FedCostWAvg: A new averaging for better Federated Learning[J]. arXiv preprint arXiv:2111.08649, 2021.
		
		\bibitem{ref_article1}
		z. iek, A. Abdulkadir, S. S. Lienkamp, T. Brox, and O. Ronneberger, 3d u-net: Learning dense volumetric segmentation from sparse annotation, in Springer, Cham, 2016.
		
		\bibitem{ref_article1}
		Gabrielsson R B. Topological Data Analysis of Convolutional Neural Networks' Weights on Images[J].
		
		\bibitem{ref_article1}
		Giannis Nikolentzos, Polykarpos Meladianos, and Michalis Vazirgiannis. Matching Node Embeddings for Graph Similarity. In Proceedings of the 31st AAAI Conference on Artificial Intelligence, 2429-2435. 2017.
		
		\bibitem{ref_article1}
		Yue Zhao, Meng Li, Liangzhen Lai, Naveen Suda, Damon Civin, and Vikas Chandra. Federated learning with non-iid data. arXiv preprint arXiv:1806.00582, 2018. 2
		
		\bibitem{ref_article1}
		Li X, JIANG M, Zhang X, et al. FedBN: Federated Learning on Non-IID Features via Local Batch Normalization[C]//lnternational Conference on Learning Representations. 2020.
		
		\bibitem{ref_article1}
		Li T, Sahu AK, Zaheer M, et al. Federated optimization in heterogeneous networks[J]. Proceedings of Machine Learning and Systems, 2020, 2: 429-450.
		
		\bibitem{ref_article1}
		Sattler F, Wiedemann S, et al. Robust and communication-efficient federated learning from non-iid data[J]. IEEE transactions on neural networks and learning systems, 2019, 31(9): 3400-3413.
		
		\bibitem{ref_article1}
		Sai Praneeth Karimireddy, Satyen Kale, Mehryar Mohri, Sashank J. Reddi, Sebastian U. Stich, and Ananda Theertha Suresh. Scaffold: Stochastic controlled averaging for federated learning. ICML, 2020. 2
		
		\bibitem{ref_article1}
		Xiangyi Chen, Tiancong Chen, Haoran Sun, Zhiwei Steven Wu, and Mingyi Hong. Distributed training with heterogeneous data: Bridging median-and mean-based algorithms. NeurIPS, 2020. 2
		
		\bibitem{ref_article1}
		Guo, P., Wang, P., Zhou, J., Jiang, S., Patel, V.M.: Multi-institutional collaborations for improving deep learning-based magnetic resonance image reconstruction using federated learning. In: Proceedings of the IEEE/CVF Conference on Computer Vision and Pattern Recognition. pp. 2423-2432 (2021)
		
		\bibitem{ref_article1}
		Guo, Pengfei, et al. Auto-FedRL: Federated Hyperparameter Optimization for Multi-institutional Medical Image Segmentation. arXiv preprint arXiv:2203.06338 (2022).
		
		\bibitem{ref_article1}
		Yeganeh Y, Farshad A, Navab N, et al. Inverse distance aggregation for federated learning with non-iid data[M]//Domain Adaptation and Representation Transfer, and Distributed and Collaborative Learning. Springer, Cham, 2020: 150-159.

		
}\end{thebibliography}
%

\end{document}